\documentclass[letterpaper]{article} % DO NOT CHANGE THIS
\usepackage{aaai24}  % DO NOT CHANGE THIS
\usepackage{times}  % DO NOT CHANGE THIS
\usepackage{helvet}  % DO NOT CHANGE THIS
\usepackage{courier}  % DO NOT CHANGE THIS
\usepackage[hyphens]{url}  % DO NOT CHANGE THIS
\usepackage{graphicx} % DO NOT CHANGE THIS
\usepackage{subcaption}
\usepackage{natbib}  % DO NOT CHANGE THIS AND DO NOT ADD ANY OPTIONS TO IT
\usepackage{caption} % DO NOT CHANGE THIS AND DO NOT ADD ANY OPTIONS TO IT
\usepackage{algorithm}
\usepackage{algorithmic}

\usepackage{newfloat}
\usepackage{listings}
\DeclareCaptionStyle{ruled}{labelfont=normalfont,labelsep=colon,strut=off} % DO NOT CHANGE THIS
\floatstyle{ruled}
\newfloat{listing}{tb}{lst}{}
\floatname{listing}{Listing}
\title{HarvestNet: A Dataset for Detecting Smallholder Farming Activity Using Harvest Piles and Remote Sensing}
\author{
    Jonathan Xu\equalcontrib \textsuperscript{\rm 2}, 
    Amna Elmustafa\equalcontrib \textsuperscript{\rm 1},
    Liya Weldegebriel\textsuperscript{\rm 1},
    Emnet Negash\textsuperscript{\rm 3,4},
    Richard Lee\textsuperscript{\rm 1},
    Chenlin Meng\textsuperscript{\rm 1}, 
    Stefano Ermon\textsuperscript{\rm 1},
    David Lobell\textsuperscript{\rm 1}
}
\affiliations{
    \textsuperscript{\rm 1} Stanford University\\
    \textsuperscript{\rm 2} University of Waterloo\\
    \textsuperscript{\rm 3} Ghent University\\
    \textsuperscript{\rm 4} Mekelle University
   \\
 
    contact@jonathanxu.com, \{amna97, liyanet\}@stanford.edu, emnet.negash@ugent.be, \\ rjlee6@stanford.edu, \{chenlin, ermon\}@cs.stanford.edu, dlobell@stanford.edu
}

\usepackage{bibentry}
\begin{document}
\pagenumbering{arabic}
\maketitle

\begin{abstract}

Small farms contribute to a large share of the productive land in developing countries. In regions such as sub-Saharan Africa, where 80\% of farms are small (under 2 ha in size), the task of mapping smallholder cropland is an important part of tracking sustainability measures such as crop productivity. However, the visually diverse and nuanced appearance of small farms has limited the effectiveness of traditional approaches to cropland mapping. Here we introduce a new approach based on the detection of harvest piles characteristic of many smallholder systems throughout the world. We present HarvestNet, a dataset for mapping the presence of farms in the Ethiopian regions of Tigray and Amhara during 2020-2023, collected using expert knowledge and satellite images, totaling 7k hand-labeled images and 2k ground-collected labels. We also benchmark a set of baselines, including SOTA models in remote sensing, with our best models having around 80\% classification performance on hand labelled data and 90\% and 98\% accuracy on ground truth data for Tigray and Amhara, respectively. We also perform a visual comparison with a widely used pre-existing coverage map and show that our model detects an extra 56,621 hectares of cropland in Tigray. We conclude that remote sensing of harvest piles can contribute to more timely and accurate cropland assessments in food insecure regions. The dataset can be accessed through \url{https://figshare.com/s/45a7b45556b90a9a11d2}, while the code for the dataset and benchmarks is publicly available at \url{https://github.com/jonxuxu/harvest-piles}

\end{abstract}

\pagenumbering{arabic}
\section{Introduction}

%Smallholder crop production is a key source of sustenance in tropical regions, especially in Sub-Saharan Africa. The presence of active farmland is one of the best indicators that an area is currently being inhabited. Such markers are critical in tracking whether people have been displaced due geopolitical issues or changes in climate. Unfortunately, for many regions, existing cropland datasets only classify by land use type, and not whether the area is actively being farmed. Moreover, many datasets are inaccurate and updated infrequently, if at all. As such, estimates about regional population and displacement tend to wary quite widely, and are not timely enough to track the immediate impacts of ongoing humanitarian issues.

%%%%%
Smallholder farming is the most common form of agriculture worldwide, supporting the livelihoods of billions of people and producing more than half of food calories \cite{samberg2016subnational,lowder2016number}. Cost effective and accurate mapping of farming activity can thus aid in monitoring food security, assessing impacts of natural and human-induced hazards, and informing agriculture extension and development policies. Yet smallholder farms are often sparse and fragmented which makes producing adequate and timely land use maps challenging, especially in resource constrained regions. Consequently, many land use datasets \cite{zanaga2022esa,brown2022dynamic,buchhorn2020copernicus} are inaccurate and updated infrequently in such regions, if at all.

\begin{figure}
    \centering
    \includegraphics[width=1\linewidth]{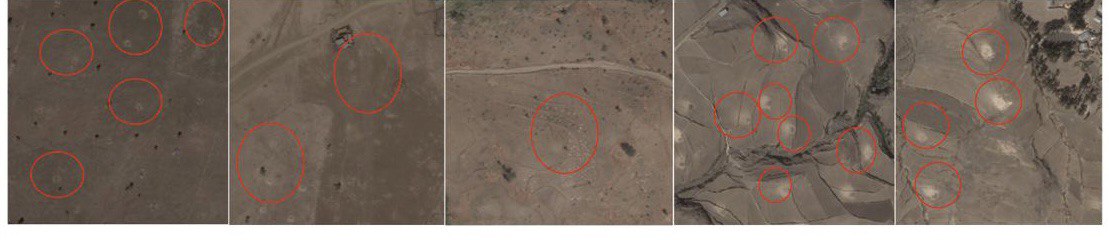}
    \caption{Various examples of harvest piles}
    \label{fig:piles}
\end{figure}
    
Machine learning algorithms for remote sensing have proved to be successful in many sustainability-related measures such as poverty mapping, vegetation and crop mapping as well as health and education measures \cite{yeh2021sustainbench}. Moreover, satellite images are now widely available at different resolutions with global coverage at low to no cost \cite{planet_scope}. The performance of methods for mapping croplands in smallholder systems, however, remains limited in many cases \cite{zanaga2022esa,brown2022dynamic}. 

Existing approaches to mapping croplands typically rely on either the unique temporal pattern of vegetation growth and senescence in crop fields compared to surrounding vegetation, the identification of field boundaries in high-resolution imagery, or some combination of both \cite{estes2022high, rufin2022large}. In non-mechanized smallholder systems like Ethiopia, where subsistence rain-fed agriculture predominates \cite{asmamaw2017critical}, these techniques face limitations. Weeds and wild vegetation often exhibit growth and spectral reflectance patterns resembling cultivated crops, causing confusion in spectral-based classification. The landscape's heterogeneity in smallholder systems, encompassing various land uses such as crops, fallow land, and natural vegetation, also poses challenges in accurately demarcating field boundaries and distinguishing different land cover types.

\begin{figure}
    \centering
    \includegraphics[width=0.9\linewidth]{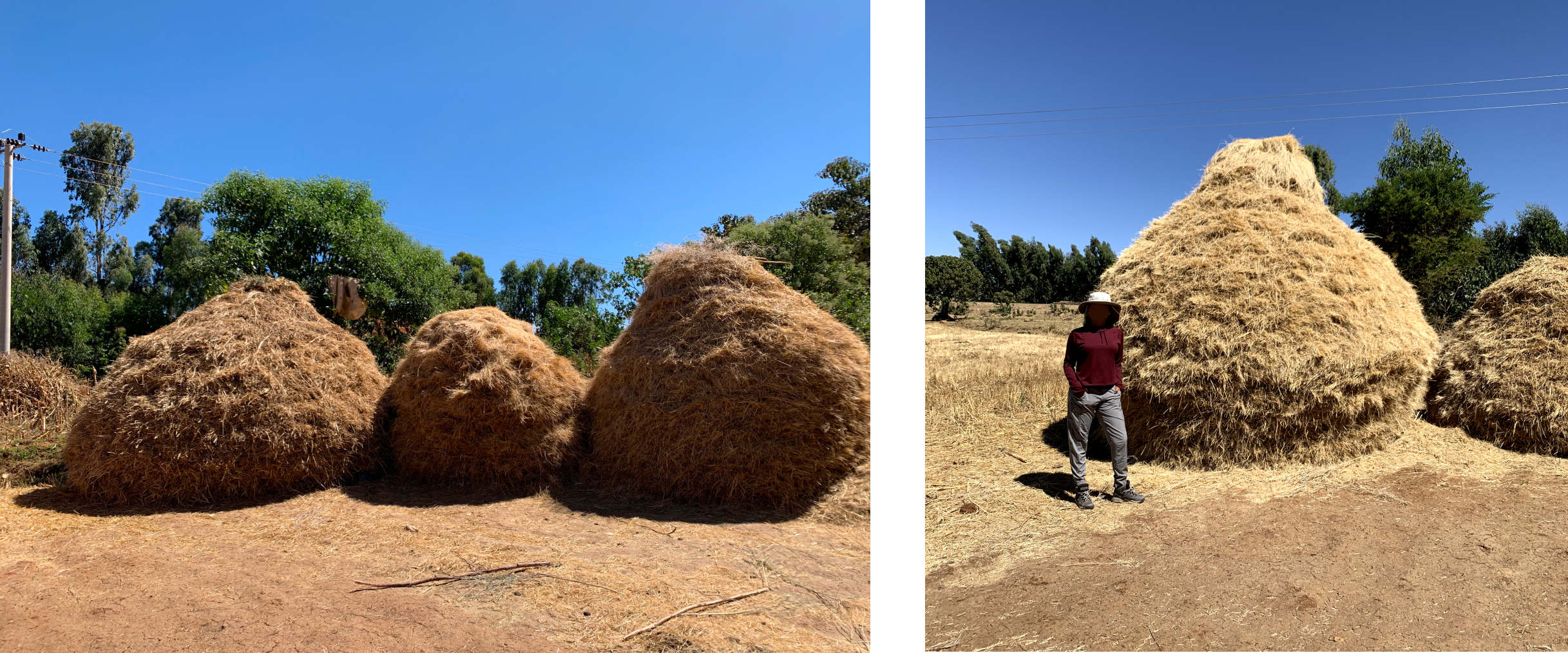}
    \caption{Photos of harvest piles. Left: person for scale.
    }
    \label{fig:natural-scale}
\end{figure}

We highlight another feature that is common in smallholder systems throughout the world — the presence of harvest piles on or near fields that cultivate grains at the end of a harvest season. Crops, particularly grains, are manually cut and gathered into piles of 3-10m before threshing, a process of separating the grain from the straw. Figure \ref{fig:natural-scale} shows what a harvest pile can look like on a natural image scale. The harvest pile footprints are present until after threshing and finally disappear when the land is prepared for the upcoming season. Since the piles are valuable, they are not abandoned in fields. Unlike houses, roads, and field boundaries, harvest piles are a more dynamic indicator that signifies seasonal farming.

We focus our work on Ethiopia, which boasts the third largest agricultural sector in Africa based on its GDP \cite{Statista}. Specifically, our attention is directed towards the lowlands in the Tigray and Amhara regions. This focus is driven by two main factors: Firstly, the area has historically been incorrectly mapped in previous works \cite{zanaga2022esa}. Secondly, this area covers arid to subhumid tropical agroclimatic zones within Ethiopia, where we have available ground data. The major crops grown in these regions include teff, barley, wheat, maize, sorghum, finger millet, and sesame \cite{ESS_report_2015, sirany2022economics}.

\begin{figure}
    \centering
    \includegraphics[width=1\linewidth]{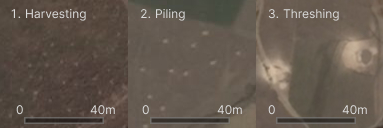}
    \caption{Various stages of harvest activity}
    \label{fig:stagesofharvest}
\end{figure}

% \se{rough transition here. maybe say something about what/how data was collected, what labels were produced, and then talk about training models. might be underselling all the work you did otherwise. basically add some text for the other contributions you listed below}
Harvest pile detection is a novel task, thus we needed to hand label our dataset to train models. To gather labels for the presence of a pile in each image, we undertook a rigorous process of hand-labeling SkySat satellite images. In this process, experts - who are researchers originally from the region and have significant field and research experience in agricultural extension work in the region - guided the identification of key areas in Tigray and Amhara. Satellite images were obtained within these areas and then AWS Mturk identified the obvious negatives while experts labeled the positives. Figure \ref{fig:piles} is a collection of various examples of piles in satellite images. In Figure \ref{fig:stagesofharvest}, we show remote sensing examples of harvest piles at various stages of harvest. We then used this labeled data to train some SOTA models in remote sensing such as CNNs and transformers and achieved 80\% accuracy on the best model. Moreover, we generated a map depicting projected farming activities in Tigray and Amhara regions, and compared it with the most current cover map.

Our contributions are as follows:
\begin{itemize}
    \item We propose a framework to detect farming activity through the presence of harvest piles.
    \item We introduce HarvestNet, a dataset of around 7k satellite images labeled by a set of experts collected for Tigray and Amhara regions of Ethiopia around the harvest season of 2020-2023.
    \item We document a multi-tiered data labeling pipeline to achieve the optimal balance of scale, quality, and consistency.
    \item We benchmarked SOTA models on HarvestNet and tested them against ground truth data and hand-labeled data to show their efficacy for the task.
    \item We produced a map for the predicted farming activity by running inference on the unlabeled data, and compared it against ESA WorldCover \cite{zanaga2022esa}, one of the most updated land usage cover map according to \cite{kerner2023accurate}. 
    
\end{itemize}

\section{Related Work}

Mapping croplands using remote sensing has been well researched in the past \cite{kussul2017deep,jiang2020large,friedl2002global,zanaga2022esa,buchhorn2020copernicus,brown2022dynamic,kerner2020rapid}. Some methods use feature engineering with nonlinear classifiers \cite{zanaga2022esa,jiang2020large,brown2022dynamic}, others use deep learning methods \cite{kerner2020rapid,kussul2017deep}. In all these works, the Normalized Difference Vegetation Index (NDVI) as well as multispectral satellite bands are used as an input, NDVI is a numerical indicator used to quantify the presence and vigor of live green vegetation by measuring the difference between the reflectance of near-infrared (NIR) and visible (red) light wavelengths in imagery. ESA \cite{zanaga2022esa} and Dynamic World \cite{brown2022dynamic} combine both NDVI and multispectral bands to provide global coverage of more than 10 classes of land use, which include crop coverage. These maps are the largest in scale and have a pixel resolution of 10m. Other methods \cite{mananze2020mapping,hackman2017new,kerner2020rapid,acharki2022planetscope} introduced a higher resolution but on a smaller scale in countries such as Mozambique, Ghana, Togo and Morocco.

Active learning is a method of building efficient training sets by iteratively improving the model performance through sampling. Some studies \cite{estes2022high, rufin2022large} have employed active learning to map smallholder farms. This approach helps mitigate bias in cropland mapping, as it can more accurately detect larger fields compared to other methods. However, none of these works have explored the concept of utilizing harvest piles as indicators when mapping smallholder farms.

\section{Method}
    In many smallholder farms for crops such as grains, farmers collect the harvest into piles during the harvest season, in preparation for threshing. These piles can be heaps of various crop types gathered around the nearest threshing ground.
    % These piles of harvest are not limited to one crop type and is common practise in many countries in Africa[].
    Therefore, the detection of piles during the harvest season is a very compelling indicator of farming activity. We propose using RGB satellite imagery for pile detection due to its wide accessibility and adaptability for other uses.
    \subsection{Task Formulation}
    \label{sec:task}
     To demonstrate this method, we defined farmland detection as a binary classification task using square RGB satellite images at a set scale. If $l$ is a location represented by latitude and longitude, the task is to build a machine learning model that takes a satellite image $x_l$ and predicts $y_l$ where $y_l$ is a binary output indicating the presence of farming activity at location $l$. The output should be positive if the image contains at least one indication of harvest activity. In our area of interest, which covers Tigray and Amhara regions in Ethiopia, the harvest process consists of three stages: cutting down and grouping crops to be collected (harvesting; Figure \ref{fig:stagesofharvest} left), piling the crops to be processed (piling; Figure \ref{fig:stagesofharvest} middle), and processing the piles to separate grains from the straw (threshing; Figure \ref{fig:stagesofharvest} right). Each stage results in different footprints of harvest patches. We classify the presence of any of these stages as a positive example of harvest activity and we use binary cross entropy loss defined by
     \begin{equation}
     L_{CE}=\frac{1}{N} \sum_l -y_l \cdot \log \left(\hat{y}_l\right)-\left(1-y_l\right) \log \left(1-\hat{y}_l\right)
    \end{equation}
    where $N$ is the number of locations $l$, $y_l$ the predictions and $\hat{y_l}$ the ground truth presence of harvest piles. More examples of harvest piles are displayed in Appendix Figure A3 and A2(see details about the Appendix in \cite{xu2023harvestnet}).

%why we choose this resolution and Time?
% we go from this task and link it to a map of active farmlands by performing inference on large scale and then create maps that indicate the availability of farmlands at some resolution. 

% \subsection{Baseline}
% %NVIDIA results from liya?
% %ESA baseline:
% We use the ESA Global Land Cover map as our target baseline for measuring our model's performance. We did this because ESA is one of the largest freely accessible land cover maps. It was created in 2020 and 2021 from images taken by Sentinel 1 and 2 satellites, and contains 11 classes of land type-one of which is the cropland class. Although ESA simply classifies whether a piece of land is cropland instead of whether it is actively being farmed, we find that the land coverage and resolution of ESA is the best for a baseline instead of (TODO). We keep this fundamental difference in classification goal during our evaluations.

\subsection{HarvestNet Dataset}
Here we introduce HarvestNet, the first dataset to our knowledge created for the task of detecting harvest activity from pile detection. Ethiopia is the second most populated country in the continent, with a majority of its people primarily dependent on smallholder rain-fed agriculture. In our regions of interest, the piling of harvests occurs during Meher, the main harvest season between September and February. These piles can be observed as early as October and stay on the land as late as May of the next year. We therefore restrict the time samples of our dataset to Oct-May months. 
% However, deciding  the scale depends on our choice of sample elevation and stage of harvest activity to look for. 
% Different stages and different elevations give different indications of the farmland area. For example, during the pile processing stage in the highlands, a farm can be located a maximum of 1 km from the spot where the pile is detected. At the same time, at some elevations, the farmland can be located only 70 m from the pile location.Since we take both highlands and lowlands into consideration, we believe that a better solution would be to use a scale that can represent different cases, more specifically this would be an image side length that is between 70 m and 1 km.
% so that we can downscale or upscale easily.
A geographical scale of around 250 m was found to be a good fit for our purposes since piles are typically located within 1km from the field plot. Our images thus cover square land areas of dimensions 256x256 m.

\subsubsection{Satellite Images} 
\begin{figure}
    \centering
    \includegraphics[width=1\linewidth]{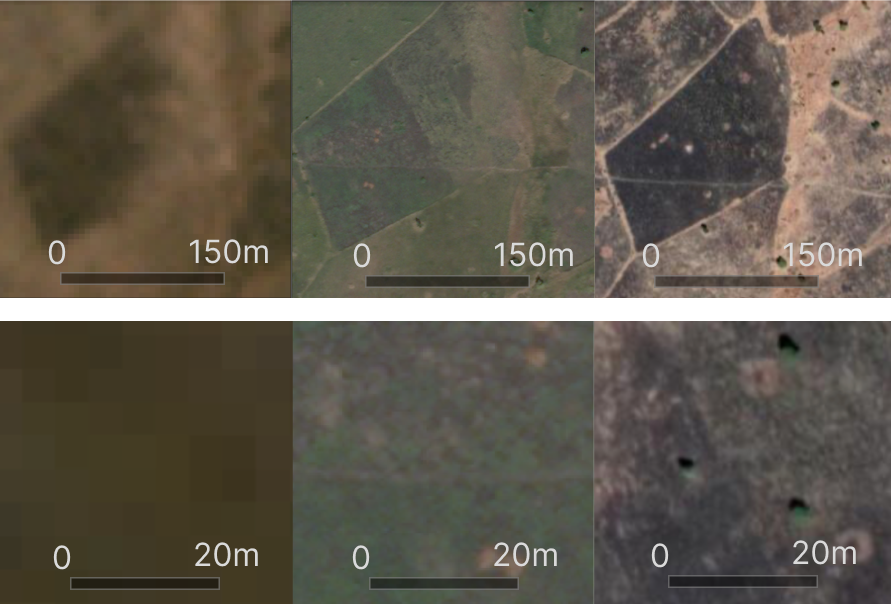}
    \caption{Side by side comparison of two areas, captured in 4.77m (left), 0.5m (center) and 0.3m (right) resolution. Note that piles become indistinguishable at 4.77m resolution.}
    \label{fig:ratio}
\end{figure}

We use two image resolutions (0.5m and 4.77m per pixel) because the small size of harvest piles necessitates high-resolution images for accurate hand labeling and mapping, as Figure \ref{fig:ratio} shows.

On the other hand, higher-res images are limited in coverage and availability. Thus, we also include around 9k (7k labeled images + 2k ground truth images) lower-res images as part of our dataset. We use the high-res images (150k unlabeled, 7k labeled images) for training and testing on the hand-labeled test set as well as for creating the crop map, while we use the low-res images for the ground truth testing since the higher res is not available in the ground truth locations. Each dataset entry includes unique latitude, longitude, altitude, and date, corresponding to SkySat images, with labeled examples also including PlanetScope images.

\textbf{SkySat} images \cite{skysat} are 512x512 pixel subsets of orthorectified composites of SkySat Collect captures at a 0.50 meter per pixel resolution. SkySat images are normalized to account for different latitudes and times of acquisition, and then sharpened and color corrected for the best visual performance. For our analysis, we downloaded every SkySat Collect with less than 10 percent cloud cover between October 2022 and January 2023. In total, we have 157k SkySat images, of which 7k are labeled. 

\textbf{PlanetScope} images \cite{planet_scope} are subsets of monthly PlanetScope Visual Basemaps with a resolution of 4.77 meters per pixel. These base maps are created using Planet Lab's proprietary "best scene on top" algorithm to select the highest quality imagery from Planet’s catalog over specified time intervals, based on cloud cover and image sharpness. The images include red, green, blue, and alpha bands. The alpha mask indicates pixels where there is no data available. We used subsets that correspond to the exact location and month of each of the 7k hand-labeled SkySat images. To maintain the same coverage of 256x256 m at the lower resolution, we used the bounding box of each SkySat image to download PlanetScope images at a size of roughly 56x56 pixels. Since the PlanetScope images are readily available and have good coverage in geography and time series, we separately downloaded 4 PlanetScope images for each area of interest corresponding to the 2k ground truth images collected by the survey team. They  include a capture for each month in the Oct-Jan harvest season. This window guarantees that farming activity will be captured in at least one of the 4 images.
\subsubsection{Labeling}
Since this is a novel task, we hand-labeled our entire training and test set. We wanted to create a high-quality, high-coverage dataset despite having limited resources and sparse access to field data and subject experts familiar with remote sensing on harvest piles. Thus, we developed a multi-staged committee approach to label successively more focused data sets. The majority of images from SkySat Collects contained no piles, so with the guidance of subject experts on agriculture in Tigray and Amhara, drew polygons outlining areas our experts knew had harvest piles (Figure 7). Downloading within those areas so, 30\% of our scenes contained harvest piles. 

The first stage of our labeling pipeline is to use crowd-sourcing to filter out obvious negatives such as images consisting only of bare lands, and shrubs. Each image is shown to 2 anonymous Amazon Mechanical Turk workers (labeler details are described in Table A6, who are each tasked with deciding whether an image contains a pile. We teach the workers a very broad definition of a "pile" so that they filter out clear negatives without accidentally discarding potential positives. When one labeler votes no and another vote yes, we (the coordinators) cast the deciding vote. Afterwards, all images labeled as positive are forwarded to two experts (co-authors of the paper) for final evaluation by their consensus. 

In Appendix Figure A4 we outline our labeling process in greater detail. The labeling process was done through inspection on SkySat images exclusively, afterwards PlanetScope images were paired with the corresponding labeled SkySat images.  By the end of this stage, we had roughly 7k labeled examples, which each consisted of a SkySat image of size 512x512 pixels and a PlanetScope image of size 56x56 pixels covering the same area at the same month.

\begin{figure}[b]
    \centering
    \includegraphics[width=1\linewidth]{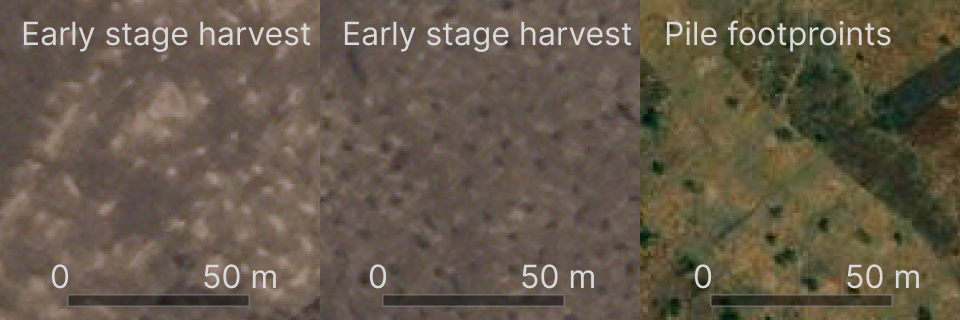}
    \caption{Examples of harvest pile activity that are not strictly piles.}
    \label{fig:positive-edgecases}
\end{figure}

\begin{figure}[ht] 
    \centering
    \includegraphics[width=1\linewidth]{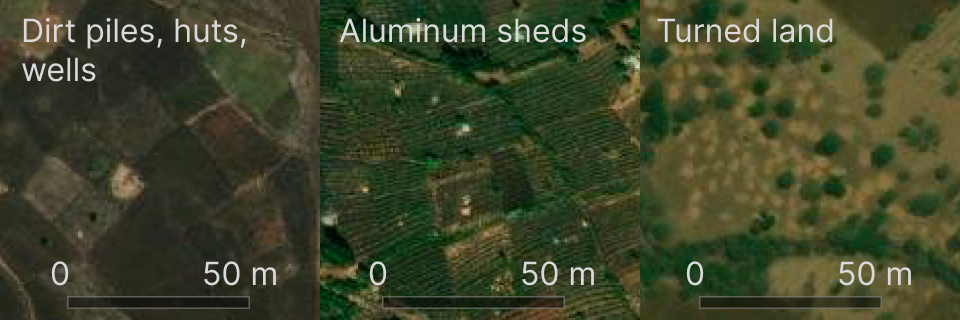}
    \caption{Examples of edge cases that are not harvest piles.}
    \label{fig:negative-edgecases}
\end{figure}

During the labeling process, we encountered diverse edge cases. Some image features resulted from the harvest piling process but did not match the conventional stage of harvest activity shown in Figure \ref{fig:piles}. Notable examples, depicted in Figure \ref{fig:positive-edgecases}, include early-stage light and dark crop bunches and residual pile footprints. These were labeled as positive instances. Additionally, some images depicted small dots resembling harvest piles, which were later identified, through consultation with our experts, as various entities such as dirt piles, aluminum sheds, and altered land shown in Figure \ref{fig:negative-edgecases}. These were deemed unrelated to harvest activity and marked as negative instances.

\subsubsection{Ground Truth}
In March 2023, we sent a survey team to collect ground truth data in Tigray and Amhara to validate our models' predictions for the 2022-2023 harvest season. 1,017 and 1,279 labels were gathered in Tigray and Amhara regions respectively. Ground truth data were gathered for all harvest crop types, including maize, teff, wheat, and finger millet. All the heaps belong to the pile point category and are situated within a maximum distance of 500 meters from the field plot. A map of ground truth collection zones is plotted in Appendix Figure A5. Due to the ongoing armed conflict, the team was unable to visit areas in Tigray that were covered by SkySat (higher-res imagery) in our image dataset. In response, we opted to combine the ground truth data with PlanetScope images, a more diverse collection that spans the geographic area with an extensive temporal range.

\subsubsection{Dataset Split}
\begin{figure}[b]
    \centering
    \includegraphics[width=1\linewidth]{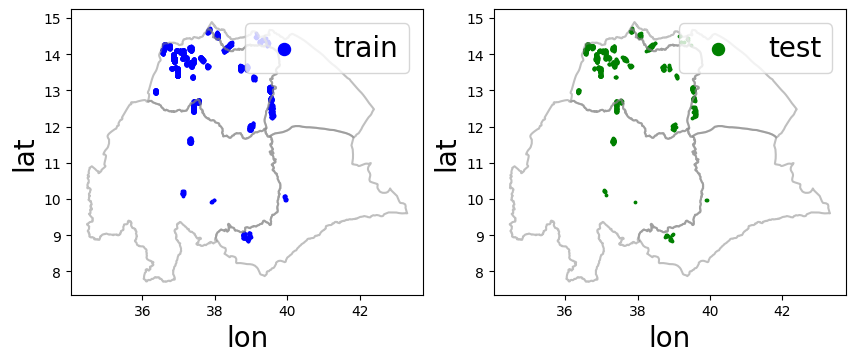}
    \caption{Train-Test split}
    \label{fig:split}
\end{figure}

Aiming for a balanced dataset, we targeted an equal split of positive and negative labels. We were able to collect SkySat images from various regions shown in Figure \ref{fig:split}, that are representative of the diversity of the geography. The exact distribution of the dataset geography and labels is described in Appendix Figure A1.

To avoid contamination from overlapping images, we used graph traversal to form distinct groups. Each group consists of images that strictly overlap with at least one other image in the group. Images that did overlap any others were assigned to their own individual groups. After graph traversal, our 6915 total images were divided into 5166 non-overlapping groups. Assigning each connected component a random color, the non-overlapping nature of groups are seen in Figure \ref{fig:contiguous} A (the code is provided in Appendix Listing A1). Afterwards, we assigned each group between the train and test split. The largest group (776 images) and the second largest (323 images) were included in the training set, because they dominate a significant area in our dataset. Then, we randomly shuffled the remaining groups, and iteratively assigned them to either the train or test set to maintain a running 80:20 ratio between the train and test sets. This results in a train/test split (Figure \ref{fig:contiguous} B) that does not overlap geographically, while still sharing a similar geographic distribution.

\begin{figure} 
    \centering
    \includegraphics[width=1\linewidth]{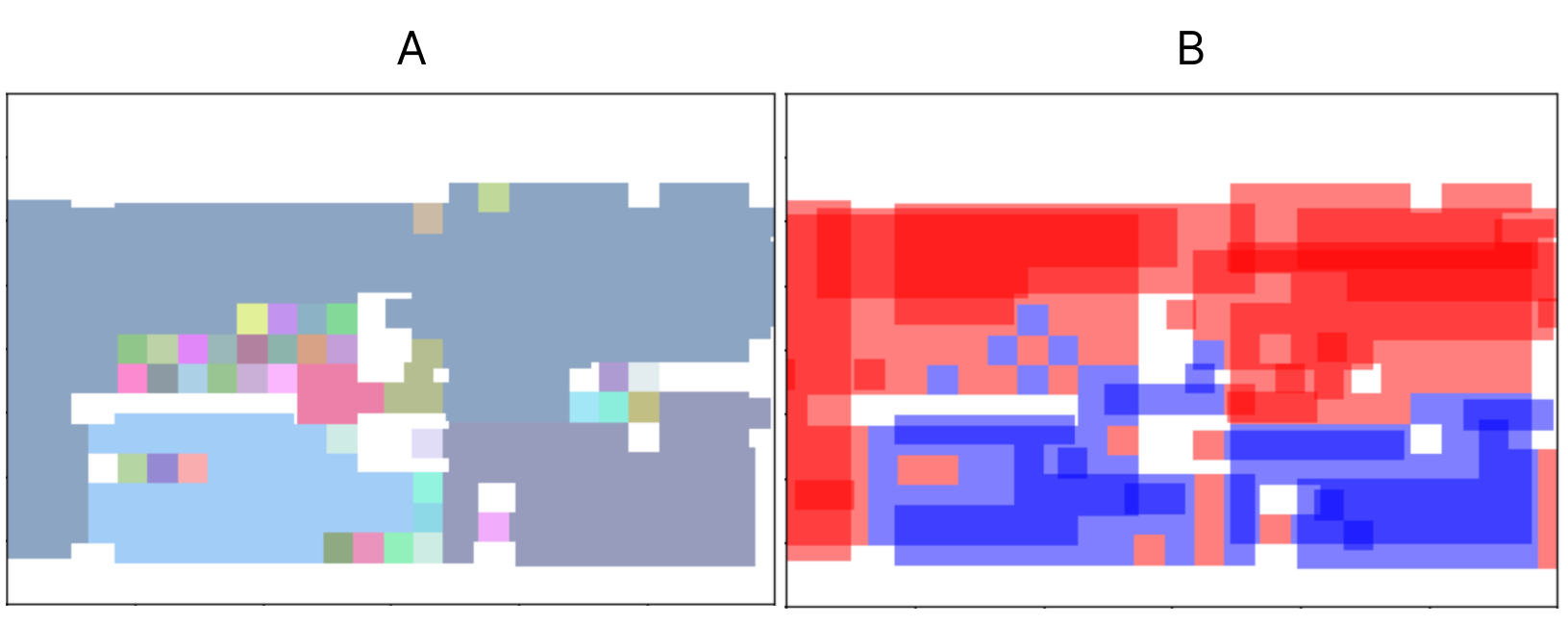}
    \caption{(A): An example region of image captures, organized into non-overlapping partitions of overlapping shapes, each assigned a random color. (B): Partitions are then divided into train (red) and test (blue).
    }

\label{fig:contiguous}
\end{figure}

\subsection{Benchmarking}
We trained various machine learning models on our dataset to predict the presence of harvest activity in an image, as described below.

\subsubsection{MOSAIKS \cite{rolf2021generalizable}}

This approach uses non-deep learning to extract features from satellite images by convolving randomly chosen patches. These features are then utilized for downstream tasks, providing cost-effective performance. We featurize our dataset with 512 features per image and employ an XGBoost classifier for target prediction.

\subsubsection{SATMAE \cite{cong2022satmae}}
Based on masked autoencoders (MAE), this framework is pretrained on FMOW and Sentinel2 for various tasks, including single image, temporal, and multispectral. It exhibits strong performance in downstream and transfer learning tasks. We apply transfer learning by training this pre-trained model to predict the harvest pile's presence in our dataset.

\subsubsection{Swin Autoencoder \cite{liu2022swin}} is a type of vision transformer that builds hierarchical feature maps by merging image patches in deeper layers and has linear computation complexity to input image size by computing self-attention only within each local window. We pretrain a masked image autoencoder built on Swin Transformer V2, using our 150k Skysat images. The input image is scaled to 224x224 pixels, and divided into a grid of patches of size 28x28. We use a mask ratio of 40\%. Then we attach a fully connected layer to the transformer's pooled output of dimension 1 x 768. The model is then fine tuned on our training set of labeled Skysat images.

\subsubsection{Satlas \cite{bastani2022satlas}}
is a pre-trained model based on the Swin transformer, and pretrained on 1.3 million remote sensing images collected from different sources. The model performs well for in-distribution and out of distribution tasks, suggesting the benefit of pretraining on a large dataset. We used the weights pretrained on higher res images, froze the model, and trained a fully connected layer on top of the pre-trained model. 

\subsubsection{ResNet-50 \cite{he2016deep}}
Convolutional Neural Networks (CNNs) have proven to perform well in several remote sensing tasks. Here, we used ResNet-50, one of the most popular and efficient networks, to predict our target. Since our input satellite image is in RGB, we used the ImageNet initialization of the network and trained a supervised binary classification task using our labeled dataset.

\section{Experiments}
\subsection{Experimental Details}
As our working dimensions are areas of size 256x256 m, we center cropped the SkySat images to 512x512 pixels and PlanetScope images to 56x56 pixels before normalizing to zero mean and unit standard deviation. These images were then scaled to fit the default input dimensions of the models.

MOSAIKS was trained with 512 features. The deep models were trained using the Adam optimizer to minimize the binary cross-entropy loss criterion. The hyperparameters on batch size, learning rate, scheduler, and training step count are described in Appendix Table A1. We experimented with combinations of hyperparameters and settled on the best performing combinations. For transformers-based models we chose the batch size that would maximise use of the 24GB of VRAM in our graphics cards. The models were trained until they converged, and the step counts were recorded.
\subsection{Evaluation}

As the task of harvest pile detection depends on the nuances of real farm activity, it is always desirable to have both a qualitative test as well as a quantitative one. We describe both evaluations below.

\subsubsection{Qualitative Evaluation}
\begin{figure}[t!]
    \centering
    \begin{subfigure}[b]{0.3\textwidth}
        \includegraphics[width=1\linewidth]{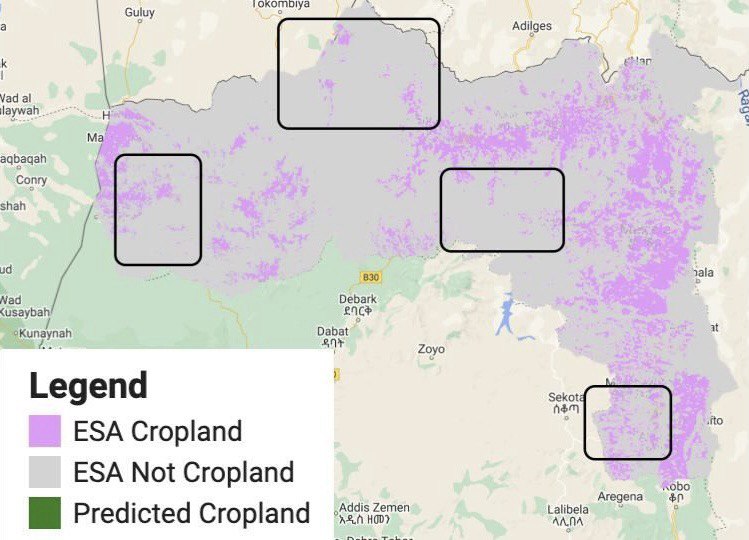} 
        \caption{}
        \label{fig:our map a}
   \end{subfigure}
  
    \begin{subfigure}[b]{0.3\textwidth}
        \includegraphics[width=1\linewidth]{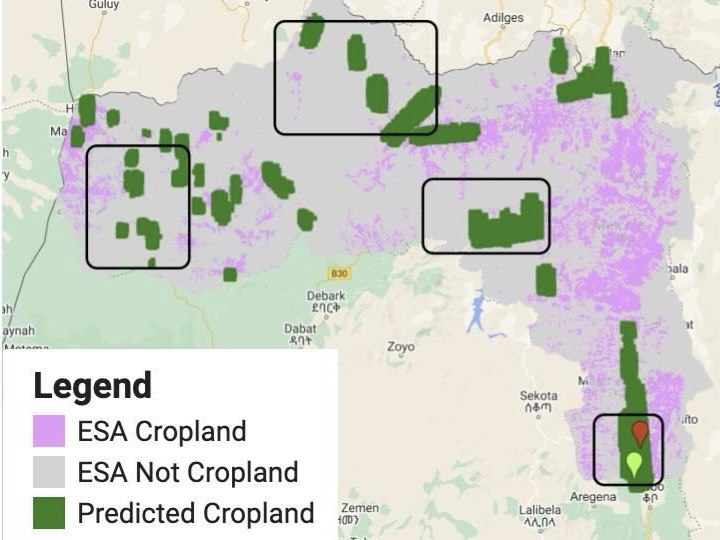}
        \caption{}
        \label{fig:our map b}
    \end{subfigure}

\begin{subfigure}[b]{0.3\textwidth}
        % \advance\leftskip-4cm
        \includegraphics[width=1\linewidth]{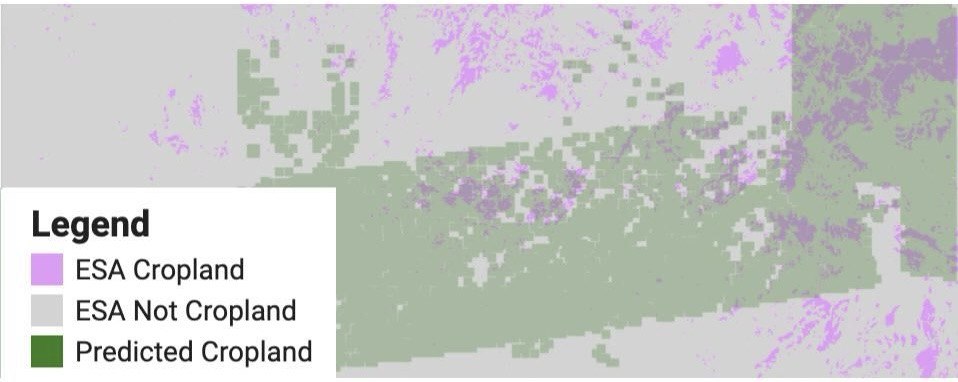}
        
        \caption{}
        \label{fig:our map c}
    \end{subfigure}
    
    \begin{subfigure}[t]{0.3\textwidth}
    \centering
        \includegraphics[height=1.2in]{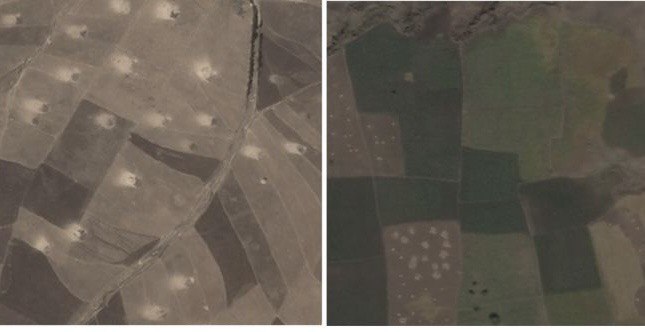}

        \caption{}
        \label{fig:our map d}
    \end{subfigure}

    \caption{(a) The ESA map for our study region. (b) Positive predictions from our ResNet-50 model, overlaid the ESA map. (c) shows a zoom-in view of (b) in the southwest of Tigray region. (d) show satellite images of the locations pinpointed in (b)}
    \label{fig:our map}
\end{figure}

We visually compare the ESA \cite{zanaga2022esa} land cover map with our ResNet-50-based classification map trained on HarvestNet. ESA is a land use map, providing global coverage for 2020 and 2021 at 10 m resolution, developed and validated based on Sentinel-1 and Sentinel-2 data. It has been independently validated with a global overall accuracy of about 75\%. Despite being SOTA in mapping land cover and land use, our experts identified many errors in smallholder systems within the area highlighted in Figure \ref{fig:our map a}. In Figure \ref{fig:our map b}, the positive classification (in green) of the best-performing model trained on our dataset overlays the ESA map (in pink), revealing our ability to detect new farmland in those regions. Figure \ref{fig:our map d} present satellite images of two example locations, verifying the presence of piles that were not detected by ESA and correctly identified in our map.

% \begin{figure}[!ht]
%     \centering
%     \begin{tabular}{@{}c@{}}
%         \includegraphics[width=1\linewidth]{AnonymousSubmission/sections/images/ESAFINAL.png} 
%         \small (a)
%         \label{fig:our map a}
%    \end{tabular}
  
%     \begin{tabular}{@{}c@{}}
%         \includegraphics[width=1\linewidth]{AnonymousSubmission/sections/images/OURMAPFINAL.png}
%         \small (b)
%         \label{fig:our map b}
%     \end{tabular}

% \begin{tabular}{@{}c@{}}
%         % \advance\leftskip-4cm
%         \includegraphics[width=1\linewidth]{AnonymousSubmission/sections/images/zoomedin_6.png}
        
%         \small (c)
%         \label{fig:our map c}
%     \end{tabular}
    
%     \begin{tabular}{@{}c@{}}   
%         \includegraphics[width=0.4\linewidth]{AnonymousSubmission/sections/images/sample-left.jpg}

%         \small (d)
%         \label{fig:our map d}
%     \end{tabular}
%     \begin{tabular}{@{}c@{}}
%         \centering
%         % \advance\leftskip-4cm
%         \includegraphics[width=0.4\linewidth]{AnonymousSubmission/sections/images/sample-right.jpg}

%         \small (e)
%         \label{fig:our map e}
%     \end{tabular}
      
%     \caption{(a) The ESA map for our study region. (b) Positive predictions from our ResNet-50 model, overlaid the ESA map. (c) shows a zoom-in view of (b) in the southwest of Tigray region. (d) and (e) show satellite images of the locations pinpointed in (b)}
%     \label{fig:our map}
% \end{figure}

\subsubsection{Quantitative Evaluation}
In this evaluation, we calculate the classification performance of our trained models using accuracy, AUROC, precision and recall. We also use the same metrics to measure the performance of our models against ground truth data.

\subsection{Results}

\begin{table*}[ht!]
    \centering
    \begin{tabular}{| l | c | c | c | c | c |} 
        \hline
        \textbf{Model} & \textbf{Accuracy} &\textbf{AUROC} &\textbf{Precision}&\textbf{Recall} &\textbf{F1-Score}\\
        \hline
        Satlas  & 67.17 & 62.47 & 80.0 & 30.61 & 44.28\\
        SatMAE &  60.0 & 56.35 & 57.37 & 29.73 & 39.17\\
        MOSAIKS & 55.46 & 51.81 & 47.65 & 23.59 & 31.56 \\
        Swin Autoencoder & \textbf{80.87} & \textbf{80.15} & 79.88 & \textbf{74.79} & \textbf{77.23} \\
        ResNet-50 & 79.18 & 77.85 & \textbf{81.4} & 67.61 & 73.87 \\
        % ResNet-50:Lowlands & 80.73 & 77.81 & 82.37 & 64.46 & 72.32 \\
        % ResNet-50:Highlands & 72.42 & 73.04 & 79.57 & 65.78 & 72.02 \\
        \hline
    \end{tabular}
    \caption[1]{Results for the proposed models on the hand labelled test set }
    \label{tab:results}
\end{table*}

\begin{table}[ht!]
    \centering
    \begin{tabular}{| l | c  | c | c | c |} 
        \hline
        \textbf{Model} & \textbf{Region}& \textbf{Accuracy} &\textbf{F1-Score} &\textbf{Recall} \\
        \hline
        ResNet-50 & Amhara &98.68 & 99.33 &  98.68 \\
        ResNet-50 &Tigray & 90.76 & 95.16 &  90.76 \\
        % Resnet50-LowerRes&amhara &  &&&\\
        % Resnet50-LowerRes&tigray &  &&& \\
        \hline
    \end{tabular}
    \caption[1]{Results for the ResNet model evaluated on the test ground truth data %\protect\footnotemark.
    }
    \label{tab:results2}
\end{table}

\begin{figure}[!ht]
    \includegraphics[width=1\linewidth]{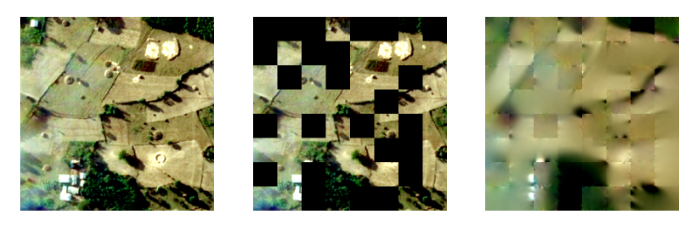}
    \caption{Reconstruction results from the Swin v2 masked autoencoder trained on 150k unlabelled Skysat imagery}
    \label{fig:reconstruction}
\end{figure}

\begin{table}[ht!]
    \centering
    \begin{tabular}{| l | c | c | c | c | c |} 
        \hline
        &\textbf{Total} & \textbf{ - , +}& \textbf{ + , +} &\textbf{ - , -} \\
        \hline
        \textbf{Samples} & 150,577 & 11,563 & 24,076  & 38,989 \\
        \textbf{Area(ha)} & 986,821 & 56,621 & 62,082 & 137,059 \\
        \hline  
    \end{tabular}
    \caption[1]{Results for the comparison between the positives and negatives predicted by the ResNet model and ESA. %\protect\footnotemark.
``-,+'' are areas where ESA predicts negative while our model predicts positive (newly detected cropland), ``+,+'' and ``-,-'' are areas where our model and ESA agree}
    \label{tab:counts}
\end{table}

Table \ref{tab:results} shows our benchmark results from the HarvestNet dataset using hand-labeled test data. Table \ref{tab:results2} presents the ResNet model's results on ground truth data. We use this model for its superior precision.

Figure \ref{fig:reconstruction} illustrates the Swin masked autoencoder's reconstruction, pretrained on 150k SkySat images. We can see that although the model was not trained on the input image, it generalizes well on filling in the masked area for Ethiopian landscapes. The model was trained on an 80\% split of the images, and evaluated on the remaining 20\% split. 

Figure \ref{fig:our map} compares the ESA map (Figure \ref{fig:our map a}) with our predicted map (Figure \ref{fig:our map b}). We highlight specific areas, marked in black rectangles, where experts identified ESA's classification inaccuracies. A closer comparison of these regions of interest can be viewed in Appendix Figure A6. Satellite images from two locations (Figure \ref{fig:our map d} ) show examples of ESA's inaccurate cropland detection. Table \ref{tab:counts} lists the differences in our and ESA's map predictions. The aim is to compare our number positive/negative predictions with those of ESA, while highlighting cases where our predictions differ. Additionally, we present the additional cropland area detected by our model (in hectares) and the overlapping cropland region shared between the two models.
\section{Discussion}

\subsubsection{Model Performance}
The outcomes presented in Table \ref{tab:results} highlight a notable trend: deep models consistently outperform non-deep models that rely on feature generators such as MOSAIKS. This disparity in performance can be attributed to the nature of our task, which involves identifying piles – intricate and compact elements within an image. The intricate nature of piles demands the capabilities of a deep network to adequately capture and detect these distinctive features. 

The second notable finding is that ResNet-50 performs quite well, even outperforming SatMAE. We believe that this is mainly due to the fact that CNNs better maintain pixel structure and generate feature maps that retain spatial information, which is a critical aspect for accurately detecting piles. Following this trend, the Swin autoencoder slightly outperforms ResNet, thanks to its incorporation of low-level details in its hierarchical feature maps. Even though Satlas is based on the Swin architecture, it performs worse than the Swin autoencoder pretrained on our unlabelled images. This suggests that although it was pretrained on a very comprehensive dataset, it is not uniquely positioned to perform in specific areas of interest.

Lastly, it is worth noting that our models' precision are notably higher than their recall, indicating a relatively high dataset quality. However, to enhance performance, further inclusion of positive samples is required, presenting a potential avenue for future research.

\subsubsection{Evaluation and Coverage}
Table \ref{tab:results2} reveals promising ResNet outcomes on ground truth data, attributed to two factors. Firstly, utilizing four distinct images per location, each representing a different month within the harvest season, enhances predictions by considering the union of results. This approach is feasible due to PlanetScope imagery's availability. Secondly, the ground truth data exclusively comprises positive samples, eliminating inherent false positives and contributing to elevated accuracy.

In Table \ref{tab:counts} and Figure \ref{fig:our map}, we demonstrate our model's improvement over ESA predictions. Figure \ref{fig:our map b}  highlights locations predicted as non-crop lands by ESA but as cropland by our model. Two samples of corresponding satellite images are shown in Figure \ref{fig:our map d}. Table \ref{tab:counts} indicates instances where our model predicts farming activity and ESA does not (11k examples, 57k ha) and cases where both models predict similarly (62k samples, 199k ha). This added cropland is estimated by experts to be 90\% true cropland, showcasing potential for improving existing maps in smallholder regions using harvest pile features. Appendix Figure A6 provides a higher zoom map for these missed locations.

\subsubsection{Potential Bias}
Labelers may be biased in their interpretation of piles. Specific areas were chosen for image downloads based on pile presence, introducing sampling bias. Models trained on our dataset may exhibit bias toward processing piles with larger shapes and colors. Geographical bias exists in ground-collected data, chosen near roads for logistical reasons.
\subsubsection{Limitations and Future Work}
Due to resource constraints and the focus on small feature classification, our dataset has limitations. Binary labels on fixed 256x256 m areas yield a lower spatial resolution, chosen for practical land coverage. Subdividing images for binary classification on smaller sections could enhance this, using subdivisions from negatively labeled images as negative labels for training.

The dataset is designed for binary classification, not object detection or semantic segmentation of harvest piles. The object detection approach could be helpful by giving information about the location, size, and density of piles. As a large part of the initial challenge was to identify areas that contain any piles at all, this next step can be applied to our existing positively labelled images. It may be a fruitful endeavor to explore using zero-shot image segmentation models such as Segment Anything \cite{kirillov2023segment} to automate the process.

In Figures \ref{fig:stagesofharvest}, \ref{fig:positive-edgecases}, and Appendix Figure A3, various image features corresponding to harvest pile activity are classified as positive examples. When expert feedback is available, classifying each feature by specific type could enhance object detection, particularly in images with multiple types of harvest activity.

Incorporating time series data for harvest pile detection is a potential next step. Faced with manpower limitations, we focused on geographical diversity in our image selection. PlanetScope imagery's performance suggests improvements with training on multiple time-series captures of the same area.

Models trained on HarvestNet indicate potential applications in other agricultural settings, like hay bale detection in North America. A harvest pile dataset may offer opportunities for transfer learning to these domains.

% Challenges:
% - low spatial resolution of Planetscope made it hard to hand label
% - reduced temporal density of PlanetScope (or skysat?) mosaics for the pre-2020 era

% Initially, we aimed to train and test our model only from images in 2019 and 2020. This was because it would correlate with the Ethiopian civil war in the Tigray region, and thus we could compare against the ground truth data of conflict incident reports from Humanitarian Data Exchange []. However, due to the challenges of SkySat coverage sparsity mentioned earlier, we expanded the time window for our labelling task to incorporate image snapshots from 2022 and 2023. We believe that the benefits of more training data in more areas outweighs the ability to compare against conflict incident reports, but believe this can be an interesting direction of exploration for the future.

% https://data.humdata.org/dataset/ucdp-data-for-ethiopia?force_layout=desktop

\section{Conclusion}
In this work, we present HarvestNet, the first dataset for detecting farming activity using remote sensing and harvest piles. HarvestNet includes a dataset for both Tigray and Amhara regions in Ethiopia, totaling 7k labelled SkySat images, and 9k labelled PlanetScope images corresponding to 2k ground truth points and the 7k labelled Skysat images.
We document the process of building the dataset, present different benchmarks results on some of the SOTA remote sensing models, and conduct land coverage analysis by comparing our predictions to ESA, a SOTA land use map. We show in our comparison that we greatly improve the current ESA map by incorporating our method of pile detection. Thus, by combining our approach with existing coverage maps like ESA, we can have a direct impact on efforts to map active smallholder farming, consequently helping to better monitor food security, assess the impacts of natural and human-induced disasters, and inform agricultural extension and development policies. 

% Link to the labels and benchmark code will be available in the supplementary material. 

\section{Acknowledgements}

This work was supported by the NASA Harvest Consortium (NASA Applied Sciences Grant No. 80NSSC17K0652, sub-award 54308-Z6059203 to DBL), NSF (\#1651565), ONR (N00014-23-1-2159), CZ Biohub and HAI. The views and conclusions contained herein are those of the authors and should not be interpreted as necessarily representing the official policies, either expressed or implied, of NASA or the U.S. Government. We are grateful to the team of data collectors from Mekelle University and Amhara Water Bureau in Ethiopia. 

\bibliography{aaai24}
\appendix
\onecolumn
\newpage
\appendix

% Reset the figure and table counters
\setcounter{figure}{0}
\setcounter{table}{0}
\setcounter{lstlisting}{0}
\renewcommand{\thefigure}{A\arabic{figure}}
\renewcommand{\thetable}{A\arabic{table}}
\renewcommand{\thelstlisting}{A\arabic{lstlisting}}

\section{Appendix}

\renewcommand{\thesubsection}{\Alph{subsection}}
\subsection{Image collection}
All SkySat images were downloaded using the Planet Python SDK. This process includes account authentication, creating a session to call Planet servers, creating an order request, and downloading the order when it's ready. For our analysis we downloaded SkySat Collects, which are approximately 50-70 SkySat Scenes and 20 x 5.9 square kilometers in size. Collects were subsetted differently based on their use case. Images used for inference were produced by subsetting entire Collects into 512 x 512 pixel sized areas. Images that were partially empty were thrown away. Unlike PlanetScope, SkySat has very limited spatial and temporal availability, limiting our choices to specific regions of Tigray and Amhara. We addressed this issue while maintaining our quota by diversely sampling areas in Tigray and Amhara. All of our images were originally stored in different folders in Google Drive based on region and time, but were later merged into one folder while still maintaining temporal and spatial information. 

\subsection{Accessing the dataset}
 The dataset is made partially accessible through this link \url{https://figshare.com/s/45a7b45556b90a9a11d2}. The labels and PlanetScope images will be shared, but unfortunately we cannot release the SkySat images due to Planet Labs' licensing requirements which would render the labels useless. Additionally, the benchmark code can be found on GitHub: \url{https://github.com/jonxuxu/harvest-piles}.

We provide the dataset in a .zip folder structured as follows:
\begin{verbatim}
Dataset
    |- planetscope_images/
    |- lables_all.csv
    |- train.csv
    |- test.csv
\end{verbatim}

\subsection{ Computational resources}
We trained our models on a single NVIDIA GeForce RTX 2080 Ti GPU with a fixed seed. MOSAIKS was trained with 3 different seeds and the average of these seeds was reported. The Swin masked autoencoder was pretrained on the task of reconstructing masked patches, and the model converged in 23 hours. The pretrained models were fine tuned for at most 5 hours. 
\subsection{Training parameters}
In Appendix Table \ref{tab:hyperparameters}, we outline the different hyperparameters of the deep models we used. Our models were all trained for 200 epochs, and the epoch count where they converged is recorded in the table. All other unlisted parameters were set to their defaults. 
\begin{table*}[ht]
    \centering
   
    \caption[1]{Hyperparameters of models trained on HarvestNet}
    \begin{tabular}{| l | c | c | c | c | c |} 
        \hline
        \textbf{Model} & \textbf{Batch size} & \textbf{Scheduler} & \textbf{Learning rate} & \textbf{Training steps}&\textbf{Convergence epochs} \\
        \hline
        Satlas & 50 & Warmup cosine & 3e-4 & 6000 & 55\\
        SatMAE & 64 & Warmup cosine & 3e-4 & 2500 & 29 \\
        Swin Autoencoder & 50 & Linear & 1e-3 & 4500 & 40\\
        ResNet-50 & 32 & One cycle & 1e-3 & 2600 & 15 \\
        \hline
    \end{tabular}
    \label{tab:hyperparameters}
\end{table*}

\subsection{ Split counts}
In Appendix Table \ref{tab:splits}, we provide counts for each train test split in both Tigray and Amhara, we also show counts of positives and negative examples in each split.
\begin{table*}[ht]
    \centering
    \caption[1]{Split counts for the train and test set, based on region and label}
    \begin{tabular}{| l | c | c | c | c | c |} 
        \hline
        &\textbf{Tigray} & \textbf{Amhara}& \textbf{Positives} &\textbf{ Negatives} &\textbf{Total} \\
        \hline
    
        \textbf{Train}&4737&795&2547 &2985&5532\\
        \textbf{Test}&1171 &212 & 608& 781 &1383\\
        \hline  
    \end{tabular}
    \label{tab:splits}
\end{table*}

\newpage

\subsection{Ablation studies}
In this section we explore the impact of various hyperparameters on the performance of models trained on HarvestNet.

\begin{table*}[ht]
    \centering
    \caption[1]{ResNet-50 Ablations}
    \begin{tabular}{| c | c | c | c | c | c | c |} 
        \hline
        \textbf{Pretrain} & \textbf{Optimizer} & \textbf{Accuracy} &\textbf{AUROC} &\textbf{Precision} &\textbf{Recall} &\textbf{F1-Score}\\
        \hline
        None & Adam & 65.05 & 69.31 & 60.00 & 60.41 & 59.34 \\
        IMAGENET1K\_V2 & Adam & 79.18 & 87.23 & 79.04 & 71.75 & 74.19 \\
        IMAGENET1K\_V2 & MADGRAD & \textbf{79.85} & \textbf{88.45} & \textbf{80.34} & \textbf{72.65} & \textbf{75.65} \\
        \hline
    \end{tabular}
    \label{tab:resent-ablations}
\end{table*}
ResNet-50 was trained using fp16 mixed precision, using the one\_cycle\_lr learning rate scheduler with a learning rate of 0.001.

\begin{table*}[ht]
    \centering
    \caption[1]{Satlas ablations}
    \begin{tabular}{| c | c | c | c | c | c | c |} 
        \hline
        \textbf{Variation} & \textbf{Accuracy} &\textbf{AUROC} &\textbf{Precision} &\textbf{Recall} &\textbf{F1-Score}\\
        \hline
        Modify pretrained output layer & 64.78 & 60.08 & \textbf{82.76} & 24.00 & 37.21 \\
        Append new output layer & \textbf{67.17} & \textbf{62.47} & 80.0 & \textbf{30.61} & \textbf{44.28} \\
        \hline
    \end{tabular}
    \label{tab:satlas ablations}
\end{table*}

\noindent
We first modified the default Satlas model by modifying its final projection layer output dimension from 1000 to 1, and appending a sigmoid layer on top.

\noindent
We then modified the default Satlas model by appending an FC layer with input dimension 1000 and output dimension 1 to the model, and appending a sigmoid layer on top. This performed better, which we believe is due to the fact that appending a layer maintains of the the latents learned in the pretrained weights.

\begin{table*}[ht]
    \centering
    \caption[1]{Swin ablations}
    \begin{tabular}{| c | c | c | c | c | c |} 
        \hline
        \textbf{Freeze pretrained} & \textbf{Accuracy} &\textbf{AUROC} &\textbf{Precision} &\textbf{Recall} &\textbf{F1-Score}\\
        \hline
        Yes & 70.10 & 68.57 & 68.28 & 57.43 & 62.37 \\
        No & \textbf{80.87} & \textbf{80.15} & \textbf{79.88} & \textbf{74.79} & \textbf{77.23} \\
        \hline
    \end{tabular}
    \label{tab:swin ablations}
\end{table*}

\subsection{Dataset distribution}
In Appendix Figure \ref{fig:distributions}, we show distributions of latitude, longitude and altitude on train, test sets as well as on the entire labelled set and unlabelled set. One notable feature of the dataset is that for each bucket in the histogram, there is a roughly equal number of positive and negative labels. Moreover, the ratio of train to test is also around 80:20 in all buckets. Most of our labelled altitude was between 500-1000m, this is because we were targeting lowlands, since previous work \cite{zanaga2022esa} had errors in lowlands in particular.

\begin{figure*}[ht] 
    \centering
    \includegraphics[width=1\textwidth]{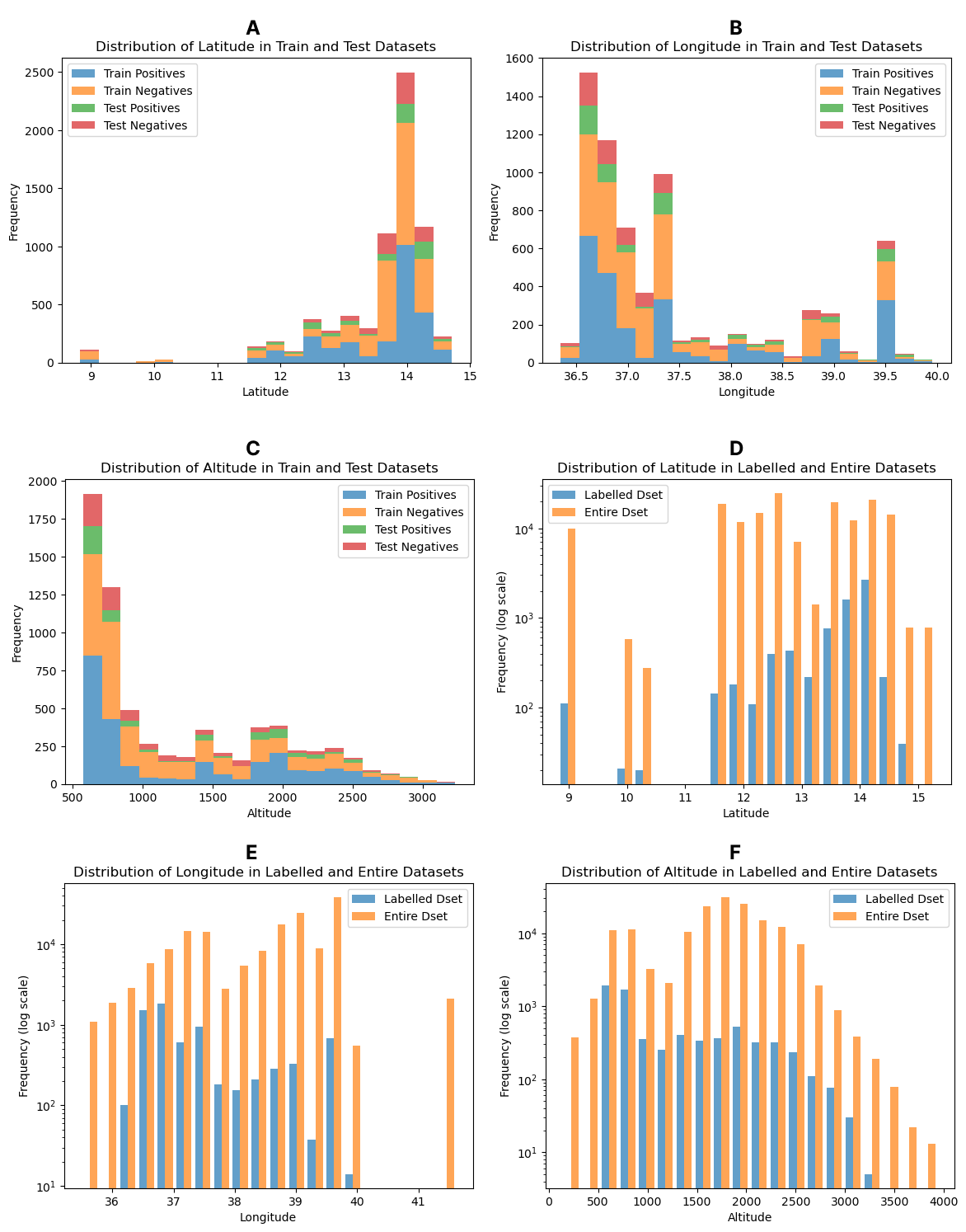}
    \caption{Statistics of HarvestNet dataset distribution}
    \label{fig:distributions}
\end{figure*}

\subsection{ Examples of harvest piles}
In Appendix Figure \ref{fig:circled-examples} and Appendix Figure \ref{fig:extra-positive-examples} we provide more examples of harvest activity. 
\begin{figure*}[ht] 
    \centering
    \includegraphics[width=1\textwidth]{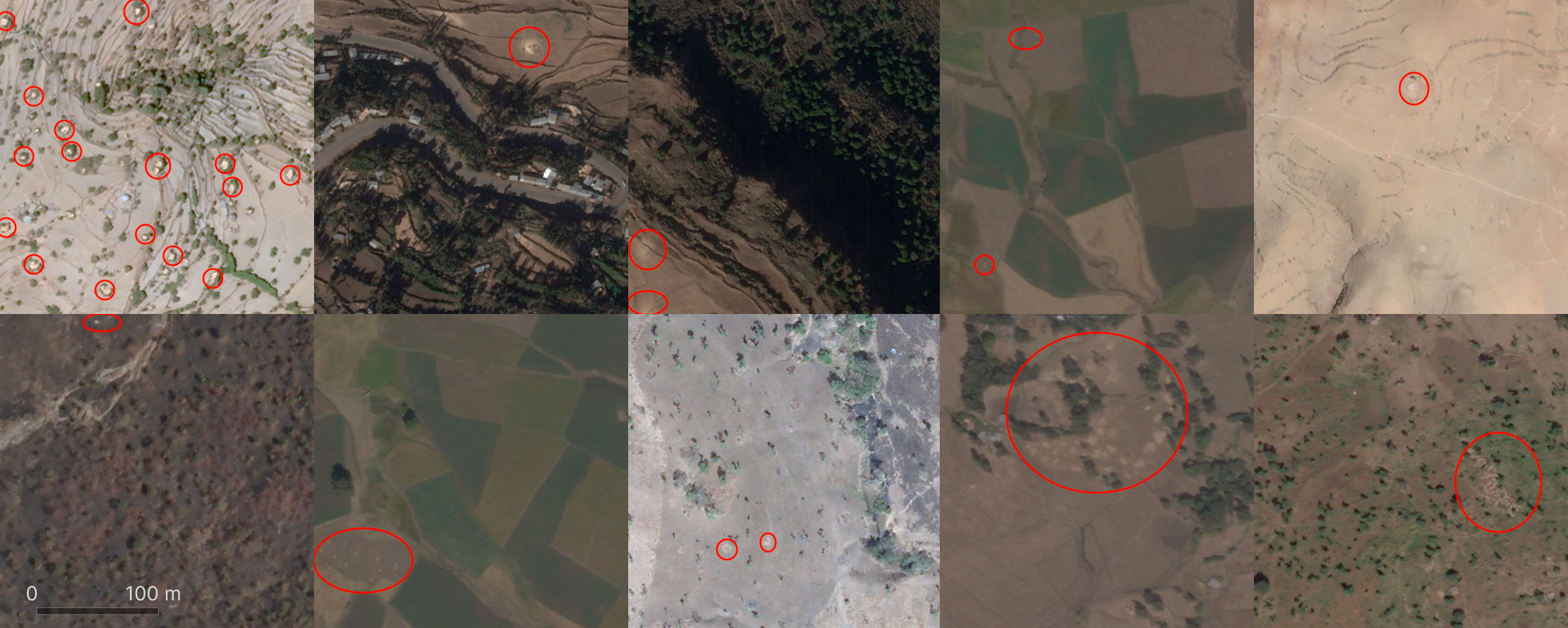}
    \caption{Examples of harvest piles at various stages, circled in red}
    \label{fig:circled-examples}
\end{figure*}
\begin{figure*}[ht] 
    \centering
    \includegraphics[width=1\textwidth]{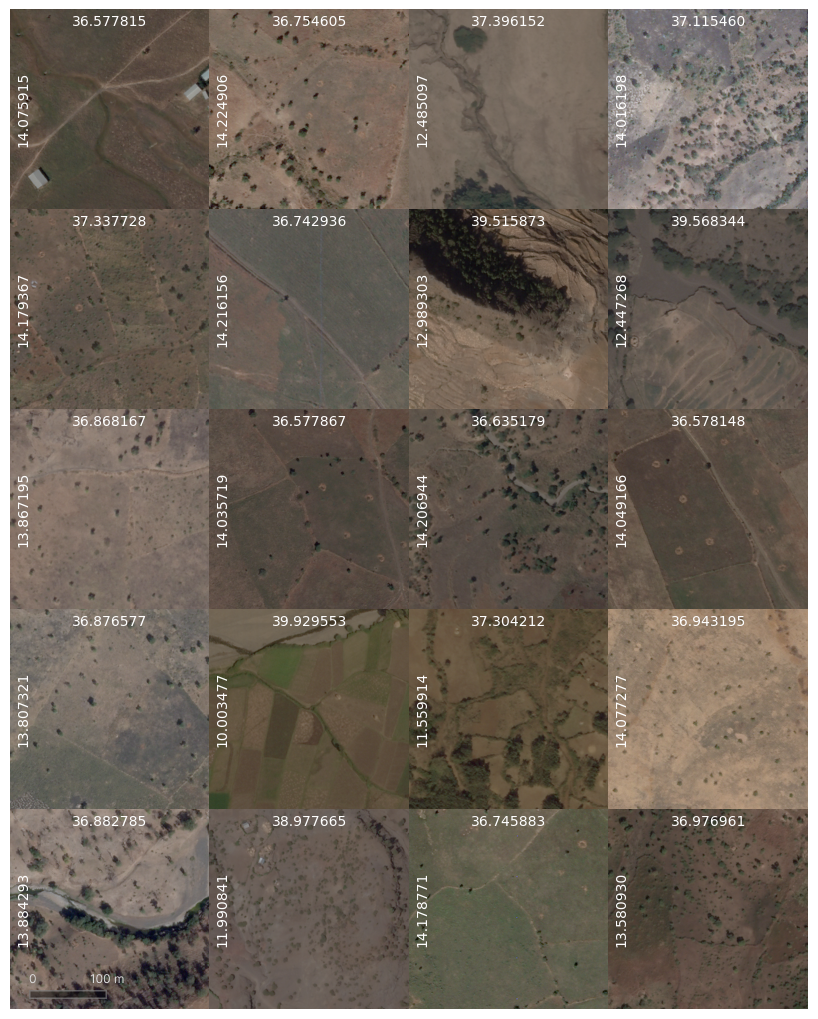}
    \caption{Additional examples of harvest pile activity, randomly selected}
    \label{fig:extra-positive-examples}
\end{figure*}

\begin{figure*}[ht] 
    \centering
    \includegraphics[width=1\textwidth]{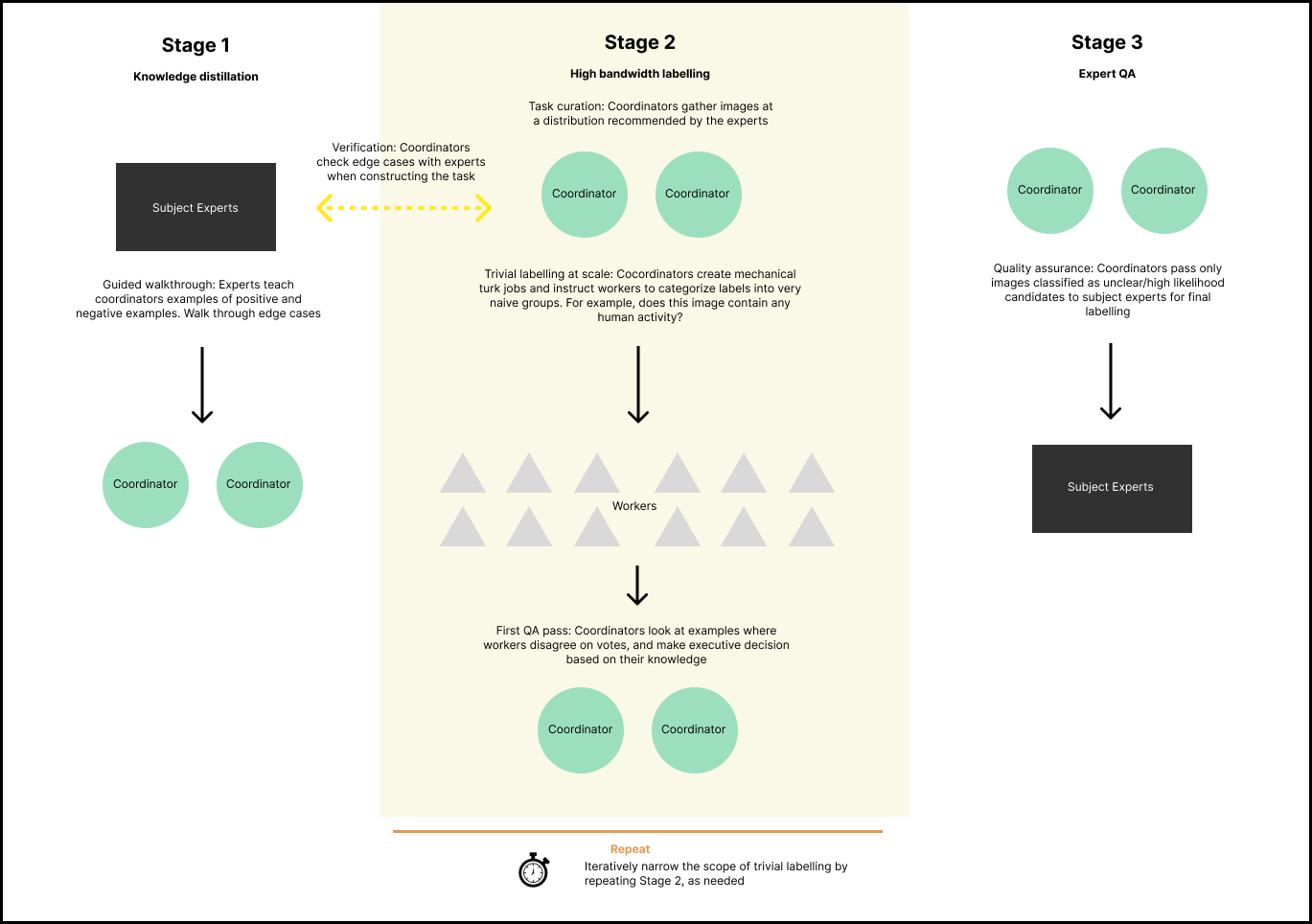}
    \caption{Labelling diagram}
    \label{fig:labelling}
\end{figure*}

\subsection{ Ground truth collection}
During February and March 2023, we sent teams of six individuals to Tigray and Amhara regions respectively to collect ground truth data. These teams had diverse backgrounds: Tigray's team included staff from Mekelle University's Department of Dryland Crop and Horticultural Sciences and Department of Land Resources Management and Environmental Protection, and staff from the College of Agriculture and Natural Resources in Mekelle, Tigray. The Amhara team was comprised of staff from the Irrigation and Lowland Area Development Bureau in Bahir Dar, Amhara.

\noindent
To gather data, the teams used handheld GPS devices, rental cars, pens, notebooks, and laptops for encoding. Guided by a map featuring available SkySat images in the 2022 harvest season, the team selected sites near roads for accessibility. Local farmers played a vital role in locating harvest pile sites. Importantly, no gathered data was discarded throughout the process. The data collection spanned about a month. 

\noindent
Both regions encountered unique challenges. In Amhara, farmer hesitation stemmed from fears of losing land to non-agricultural industries. There was also a prevailing distrust regarding the purpose of the collected data, given the significance of harvest piles for livelihoods.

\noindent
Tigray presented a unique set of challenges. Many of the chosen sites had been active battlefronts in recent years, carrying high risks of unexploded bombs. Additionally, the team faced instances of dog attacks, particularly prevalent in the Central zone where dogs had not received vaccinations for approximately two years due to the conflict. Since the troops had not yet left Tigray territory, the team faced exposure to troops from Amhara and Eritrea. There were also snake attacks in areas like the Menji-Guya line. The security situation was precarious and frightening during the field work.

\begin{figure*}[ht] 
    \centering
    \includegraphics[width=1\textwidth]{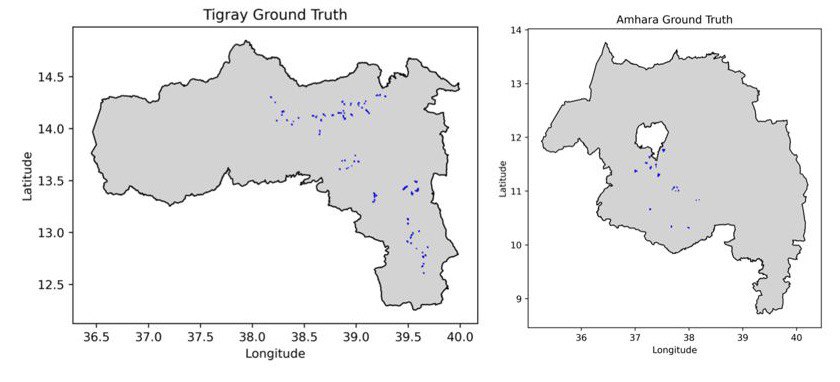}
    \caption{Ground truth collection zones in Tigray(left) and Amhara(right)}
    \label{fig:gt}
\end{figure*}

\noindent
Appendix Figure \ref{fig:gt} illustrates the geographical distribution of the 2,296 data collection points acquired by our survey team. These points span across the Tigray and Amhara regions.

\subsection{I. Closeup of ESA comparison}
\begin{figure*}[ht]
    \centering
    \begin{subfigure}[b]{1\linewidth}
        \centering
        % \advance\leftskip-4cm
        \includegraphics[width=0.6\linewidth]{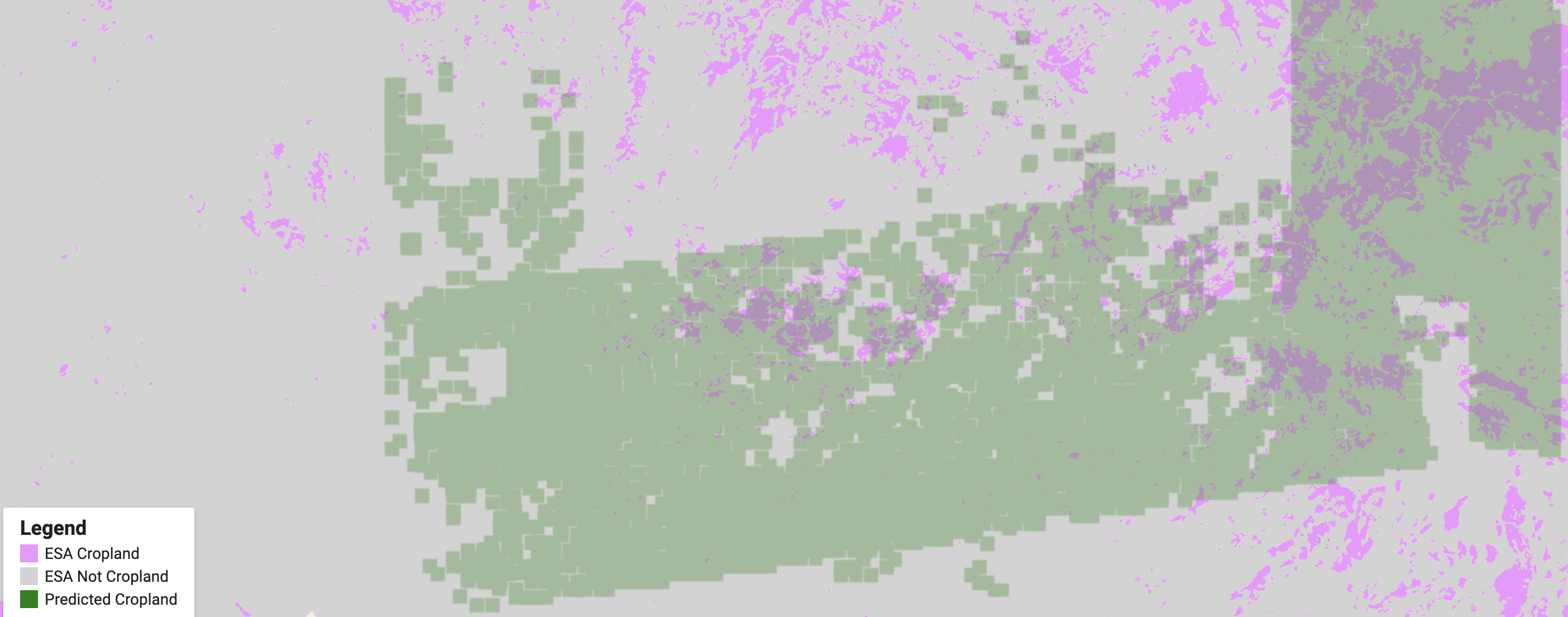}
        \caption{}
        \label{fig:zoom1}
    \end{subfigure}

    \begin{subfigure}[b]{1\linewidth}
      \centering
        \includegraphics[width=0.6\linewidth]{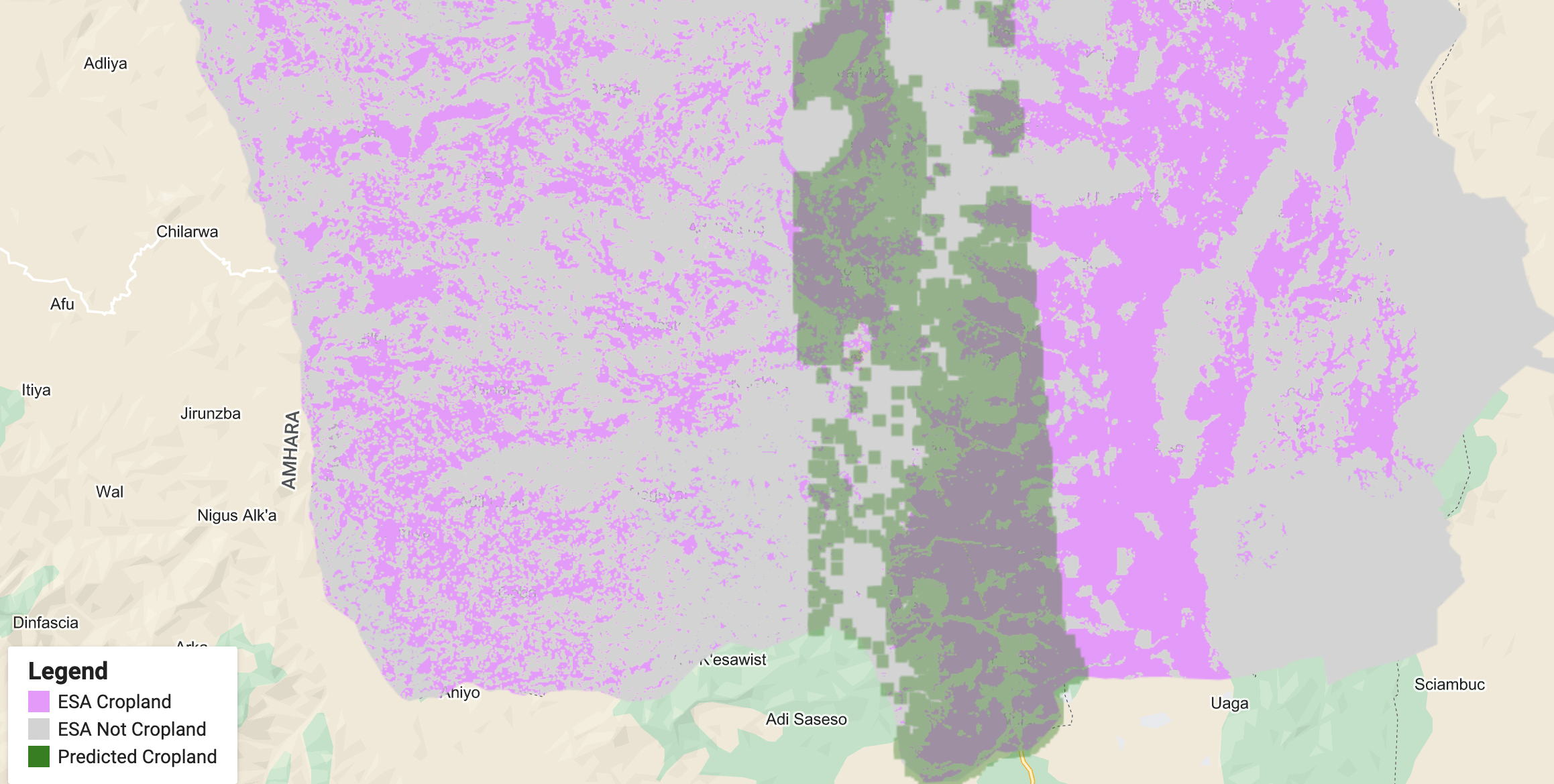}
        \caption{}
        \label{fig:zoom2}
    \end{subfigure}

    \begin{subfigure}[b]{1\linewidth}
        \centering
        \includegraphics[width=0.6\linewidth]{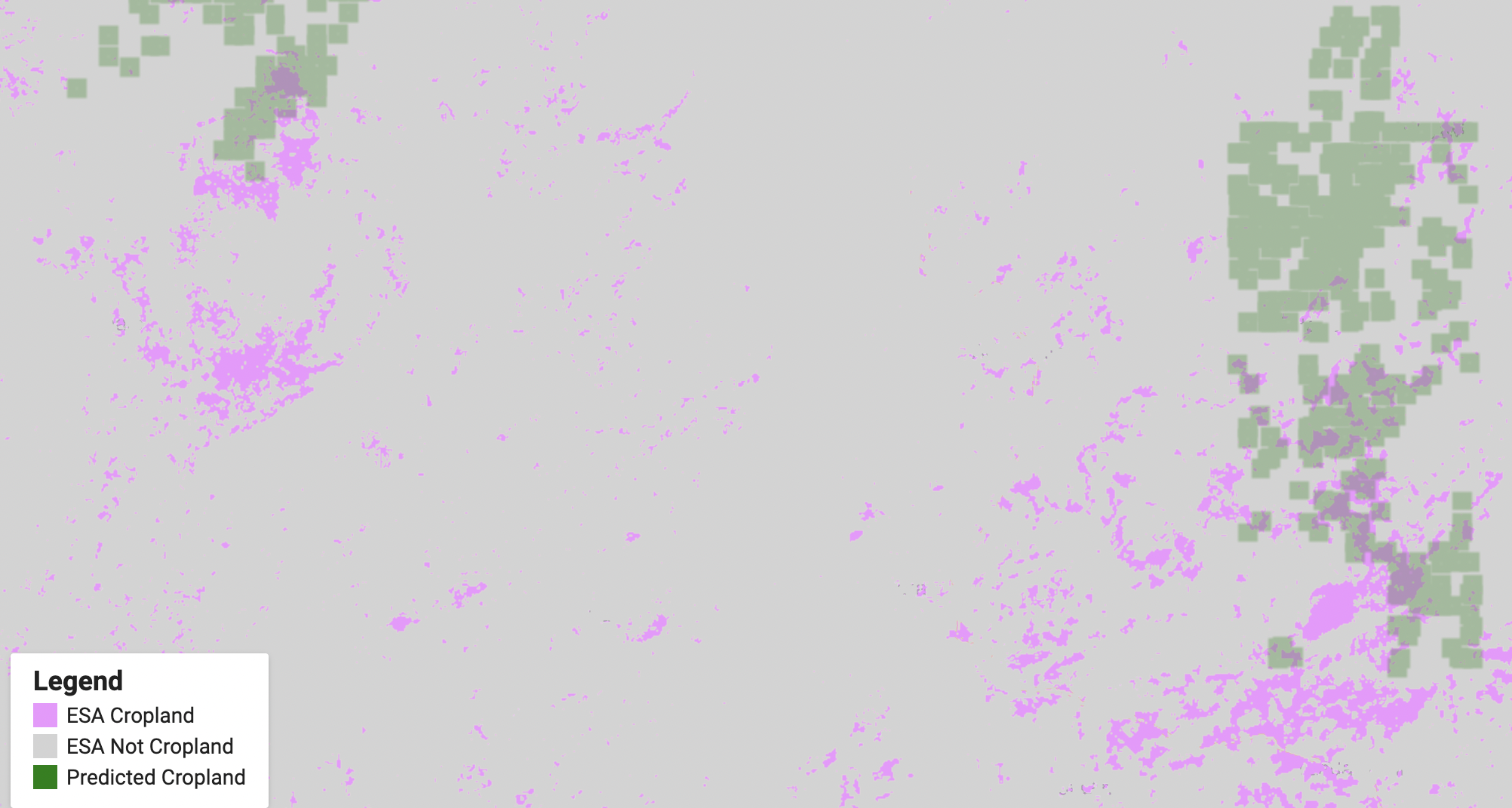}
        \caption{}
        \label{fig:zoom3}
    \end{subfigure}

    \begin{subfigure}[b]{1\linewidth}
        \centering
        \includegraphics[width=0.6\linewidth]{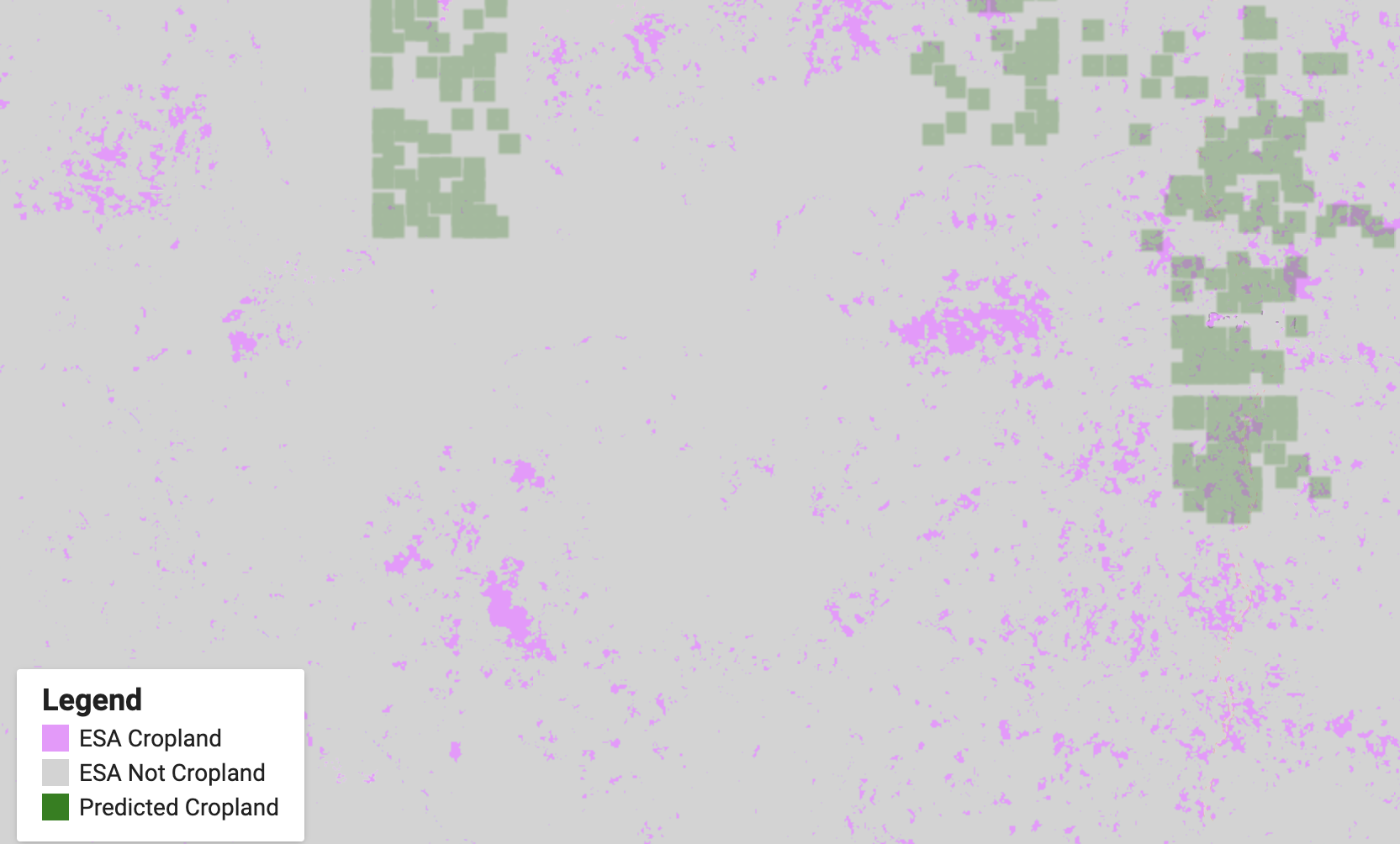}
        \caption{}
        \label{fig:zoom4}
    \end{subfigure}
    
    \caption{Close up view of ResNet-50 model predictions overlaid on top of ESA map}
    \label{fig:esa-closup}
\end{figure*}

In Figure \ref{fig:esa-closup} we show close up examples of the  locations in squares shown in Figure \ref{fig:our map a}, overlaying the ESA map in pink.

To accurately determine the additional cropland area projected by our model, we employed a systematic process. Surrounding each prediction point generated by our model, we established bounding boxes measuring 256x256 meters. Within these boxes, we evaluated the extent of coverage by the ESA cropmask, specifically targeting positive bounding boxes. If a given box had an ESA cropmask coverage of 20 percent or less, we classified it as newly predicted cropland by our model. For the shared cropland area recognized by both our model and ESA, we summed the areas of positive squares exhibiting an 80 percent or higher overlap with the ESA cropmask. Employing a similar methodology, we identified non-cropland areas mutually disregarded by both our model and ESA, by tallying the area of negative squares with an ESA cropmask coverage of 20 percent or lower.

\subsection{Partition assignment code}
\begin{lstlisting}[language=Python, caption={Contiguous shape group partitioning algorithm}, label=lst:partition]
# Create a graph with rectangles as nodes and overlaps as edges
import pandas as pd
import os
import networkx as nx
from shapely.geometry import box
from shapely.strtree import STRtree

df = pd.read_csv(os.path.join(FOLDER_PATH, "merged_labelled.csv"))
df = df.iloc[:, 1:]

G = nx.Graph()

# Create shapes and nodes
def create_rectangle(row):
    return box(row['lat_2'], row['lon_1'], row['lat_1'], row['lon_2'])
    
geometry=[]
for index, row in df.iterrows():
    G.add_node(index)
    geometry.append(create_rectangle(row))

tree = STRtree(geometry)

# Add edges for each overlapping box
for idx, shape in enumerate(geometry):
    for intersecting in tree.query(shape):
        if not shape.touches(geometry[intersecting]) and idx != intersecting:
            G.add_edge(idx, intersecting)

connected_components = list(nx.connected_components(G))
groups_of_rectangles = [list(component) for component in connected_components]
\end{lstlisting}

\subsection{Labelling procedure}
We conducted a labeling procedure with the primary objective of optimizing accuracy and leveraging expert knowledge, while simultaneously expanding the scale of our labeled dataset.  In Stage 1: knowledge distillation (Appendix Figure A4), we (coauthors) did a walkthrough of some examples guided by experts to familiarize ourselves with the appearance distribution of positive and negative examples of harvest piles. In Stage 2: high bandwidth labeling we focused on transferring a foundational proficiency to teach public labellers how to detect trivial examples of harvest activity. To achieve this, we instructed labellers by presenting multiple illustrations depicting harvest-related activities highlighted in red circles, of the same composition as shown in Appendix Figure A2. The illustrative samples were intentionally broad in classifying harvest piles; for instance, even strictly negative cases such as plastic tarps concealing sesame and accumulations of harvest remnants repurposed as animal feed were presented as affirmative instances of harvest piles. This inclusive approach was done to minimize false negative labels.

\noindent
In Stage 2 we used public labellers to relabel  3792 negative examples that were previously labelled by coordinators but denoted by experts to have many false negatives. To promote dataset quality while minimizing costs, each image was presented to two labellers, who gave a binary label after reading the instructions. Details about the batch job are listed in Appendix Table A4. We chose to increase the quality of our workers by setting minimum requirements for their historic task approval rate and count.

\noindent
It is interesting to note that our entire batch job was completed within 4 hours and 45 minutes. The efficiency of MTurk's crowd-sourced labeling capacity  open the prospects of automated quality control in significantly enhancing our labeling throughput.

\begin{table*}[ht]
    \centering
    \caption{Labelling Job Details}
    \begin{tabular}{|c|c|c|c|}
        \hline
        \multicolumn{2}{|c|}{Task details} & \multicolumn{2}{c|}{Job completion status} \\
        \hline
        Reward per assignment & \$0.01 & Assignments completed & 7584 \\
        Number of assignments per task 2 & 2 & Average time per assignment & 8 min 24 sec \\
        Time allotted per assignment & 1 hour & Creation time & June 30, 2023 9:56 AM PDT \\
        Task expires in & 2 days & Completion time & June 30, 2023 2:40 PM PDT \\
        Auto-approve and pay workers in & 3 days &  & \\
        \hline
        \multicolumn{2}{|c|}{Worker Requirements} & \multicolumn{2}{c|}{Cost summary} \\
        \hline
        Require workers to be masters & No & Total reward & \$75.84 \\
        HIT approval rate \% & Greater than 98 & Fees to Mechanical Turk & \$75.84 \\
        Number of HITs Approved & Greater than 50 & Total cost &  \$151.68 \\
        \hline
    \end{tabular}
    \label{tab:turk-details}
\end{table*}

\noindent
By the end of the crowdsourced labelling step, we had 3792 SkySat images, each labelled by two labellers. For 437 of the images, the labellers both agreed the image did not contain piles. For 1708 of the images, the labellers agreed the image contained piles. For the remaining 1647 images where the labellers did not agree, we (the coauthors and project coordinators) manually labelled the images again, using our better knowledge of the appearance of harvest piles on SkySat images. After our manual pass through, we had 1997 positively labelled images and 1795 negatively labelled images.

\noindent
The 1997 positively labelled images were then sent to Stage 3: Expert QA. Here, our subject experts manually reviewed each image that we decided were highly probable candidates for positive examples of harvest piles. After review, 341 of the 1997 images were labelled as positives, and the remaining were labelled as negatives. When we combined these updated labels with our dataset, we ended up with our current labelled dataset of 2547 positives and 2985 negatives.
\end{document}